\pdfoutput=1
\documentclass[12pt]{article}
\usepackage[hidelinks]{hyperref}
\usepackage{url}

\usepackage[round]{natbib}
\usepackage{amsmath,dsfont,graphicx,color,amssymb,amsthm}
\usepackage{algorithm,algorithmic,caption}
\usepackage{multirow} 
\usepackage{bbm}
\usepackage{soul}
\usepackage{xcolor,colortbl}
\usepackage{setspace}

\renewcommand{\eqref}[1]{Eq.~(\ref{#1})}
\newcommand{\figref}[1]{Fig.~\ref{#1}}

\newcommand{\secref}[1]{Section \ref{#1}}
\newcommand{\appref}[1]{Appendix \ref{#1}}

\newcommand{\localpoly}{{\cal{M}}_L}

\DeclareMathOperator*{\argmax}{argmax}
\newcommand{\expect}{\mathbb{E}}
\renewcommand{\cite}{\citep}

\newcommand{\reals}{\mathbb{R}}
\newcommand{\half}{\frac{1}{2}}

\newtheorem{theorem}{Theorem}[section]

\newtheorem{proposition}[theorem]{Proposition}

\newenvironment{definition}[1][Definition]{\begin{trivlist}
\item[\hskip \labelsep {\bfseries #1}]}{\end{trivlist}}

\makeatletter
\usepackage{xspace}
\def\@onedot{\ifx\@let@token.\else.\null\fi\xspace}
\DeclareRobustCommand\onedot{\futurelet\@let@token\@onedot}

\newcommand*\samethanks[1][\value{footnote}]{\footnotemark[#1]}

\makeatletter
\renewcommand*{\@fnsymbol}[1]{\ensuremath{\ifcase#1\or \dagger\or \ddagger\or
   \mathsection\or \mathparagraph\or \|\or **\or \dagger\dagger
   \or \ddagger\ddagger \else\@ctrerr\fi}}
\makeatother

\newcommand{\thmref}[1]{Theorem~\ref{#1}}

\makeatother

\renewcommand{\cite}{\citep}

\title{Train and Test Tightness of LP Relaxations in Structured Prediction}

\author{\hspace{-17pt} Ofer Meshi\thanks{Toyota Technological Institute at Chicago}
~~~Mehrdad Mahdavi\samethanks 
~~~Adrian Weller\thanks{University of Cambridge} 
~~~David Sontag\thanks{New York University} 
}

\date{}

\begin{document} 

\maketitle

\begin{abstract}
Structured prediction is used in areas such as computer vision and natural language processing to predict structured outputs such as segmentations or parse trees. In these settings, prediction is performed by MAP inference or, equivalently, by solving an integer linear program. Because of the complex scoring functions required to obtain accurate predictions, both learning and inference typically require the use of approximate solvers.
%
We propose a theoretical explanation to the striking observation that approximations based on linear programming (LP) relaxations are often tight on real-world instances.
In particular, we show that learning with LP relaxed inference encourages integrality of training instances, and that tightness generalizes from train to test data.

%
\end{abstract}

\section{Introduction}
Many applications of machine learning can be formulated as prediction problems over structured output spaces \cite{struct_pred_book,advanced_pred14}.
In such problems output variables are predicted \emph{jointly} in order to take into account mutual dependencies between them, such as high-order correlations or structural constraints (e.g., matchings or spanning trees).
Unfortunately, the improved expressive power of these models comes at a computational cost, and indeed, exact prediction and learning become NP-hard in general.
Despite this worst-case intractability, efficient approximations often achieve very good performance in practice.
In particular, one type of approximation which has proved effective in many applications is based on \emph{linear programming (LP) relaxation}.
In this approach the prediction problem is first cast as an integer LP (ILP), and then the integrality constraints are relaxed to obtain a tractable program.
In addition to achieving high prediction accuracy, it has been observed that LP relaxations are often \emph{tight} in practice.
That is, the solution to the relaxed program happens to be optimal for the original hard problem (an \emph{integral} solution is found).
This is particularly surprising since the LPs have complex scoring functions that are not constrained to be from any tractable family. 
A major open question is to understand why these real-world instances behave so differently from the theoretical worst case.

This paper aims to address this question and to provide a theoretical explanation for the tightness of LP relaxations in the context of structured prediction.
In particular, we show that the approximate training objective, although designed to produce accurate predictors, also induces tightness of the LP relaxation as a byproduct.
Our analysis also suggests that exact training may have the opposite effect. 
To explain tightness of \emph{test} instances, we prove a generalization bound for tightness. Our bound implies that if many training instances are integral, then test instances are also likely to be integral. Our results are consistent with previous empirical findings, and to our knowledge provide the first theoretical justification for the wide-spread success of LP relaxations for structured prediction in settings where the training data is not linearly separable.
\section{Related Work}
Many structured prediction problems can be represented as ILPs \cite{roth2005integer,martinsACL09,RusSonColJaa_emnlp10}. 
Despite being NP-hard in general \cite{Roth96,Shimony94}, various effective approximations have been proposed.
Those include both search-based methods \cite{searn,zhang2014greed}, and natural LP relaxations to the hard ILP \cite{Schlesinger,koster,chekuri2004linear,martinLP}.
Tightness of LP relaxations for special classes of problems has been studied extensively in recent years and include restricting either the structure of the model or its score function.
For example, the pairwise LP relaxation is known to be tight for tree-structured models and for supermodular scores \citep[see, e.g.,][]{jordan, ThapperZivnyFOCS12},
and the cycle relaxation (equivalently, the second-level of the Sherali-Adams hierarchy) is known to be tight both for planar Ising models with no external field \cite{barahonaplanar} and for almost balanced models \cite{weller16}. 
To facilitate efficient prediction, one could restrict the model class to be tractable. 
For example, \citet{Taskar04} learn supermodular scores, and \citet{mtreen} learn tree structures.



\begin{figure}
 \begin{center}
   \includegraphics[width=2.3in]{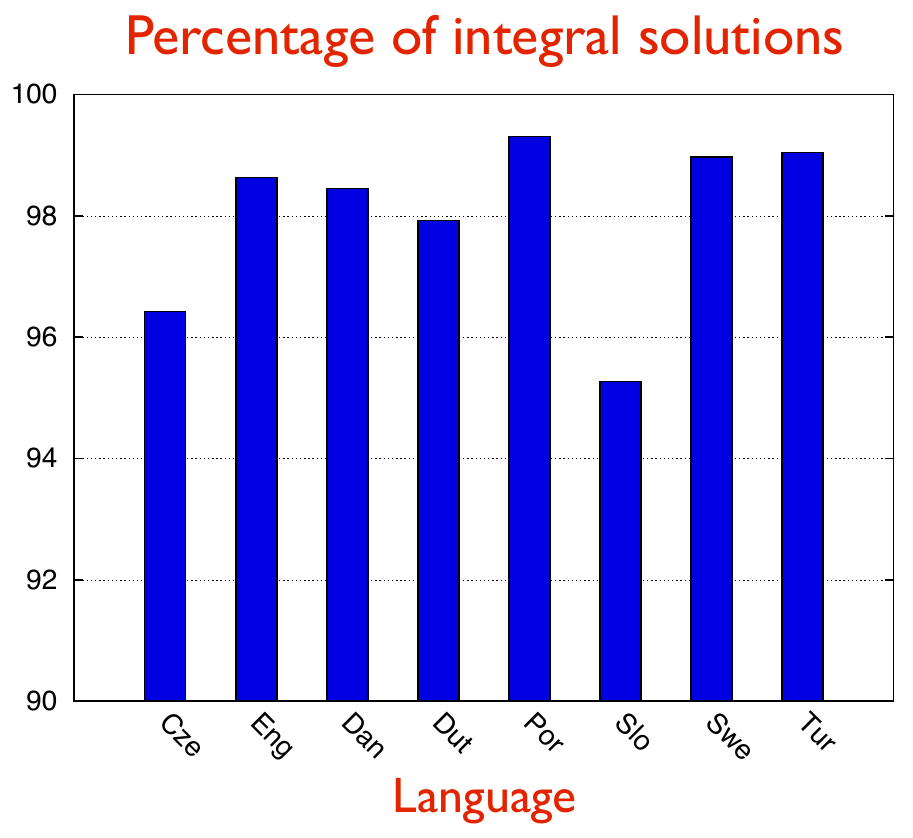}
 \end{center}
\vspace{-4mm}
 \caption{Percentage of integral solutions for dependency parsing from \citet{KooEtAl10}.}
  \label{fig:parsing}
  \vspace{-0.4cm}
\end{figure}

However, the sufficient conditions mentioned above are by no means necessary, and indeed, many score functions that are useful in practice do not satisfy them but still produce integral solutions \cite{RothYih04,SontagEtAl08,Finley08,martinsICML09,KooEtAl10}.
For example, \citet{martinsICML09} showed that predictors that are learned with LP relaxation yield integral LPs on $92.88\%$ of the test data on a dependency parsing problem (see Table 2 therein). \citet{KooEtAl10} observed a similar behavior for dependency parsing on a number of languages, as can be seen in \figref{fig:parsing} (kindly provided by the authors).
The same phenomenon has been observed for a multi-label classification task, where test integrality reached $100\%$ \citep[][Table 3]{Finley08}.

Learning structured output predictors from labeled data was proposed in various forms by \citet{Collins02,Taskar03,Tsochantaridis04}.
These formulations generalize training methods for binary classifiers, such as the Perceptron algorithm and support vector machines (SVMs), to the case of structured outputs.
The learning algorithms repeatedly perform prediction, necessitating the use of approximate inference within training as well as at test time.
A common approach, introduced right at the inception of structured SVMs by \citet{Taskar03}, is to use LP relaxations for this purpose.

The most closely related work to ours is \citet{Kulesza07}, which showed that not all approximations are equally good, and that it is important to match the inference algorithms used at train and test time. The authors defined the concept of {\em algorithmic separability} which refers to the setting when an approximate inference algorithm achieves zero loss on a data set. The authors studied the use of LP relaxations for structured learning, giving generalization bounds for the true risk of LP-based prediction. However, since the generalization bounds in \citet{Kulesza07} are focused on prediction {\em accuracy}, the only settings in which tightness on test instances can be guaranteed are when the training data is algorithmically separable, which is seldom the case in real-world structured prediction tasks (the models are far from perfect). Our paper's main result (Theorem~\ref{thm:integrality_generalization_bound}), on the other hand, guarantees that the expected fraction of test instances for which a LP relaxation is {\em integral} is close to that which was estimated on training data. This then allows us to talk about the generalization of {\em computation}. For example, suppose one uses LP relaxation-based algorithms that iteratively tighten the relaxation, such as \citet{sontag, SontagEtAl08}, and observes that 20\% of the instances in the training data are integral using the pairwise relaxation and that after tightening using cycle constraints the remaining 80\% are now integral too. Our generalization bound then guarantees that approximately the same ratio will hold at test time (assuming sufficient training data).

\citet{Finley08} also studied the effect of various approximate inference methods in the context of structured prediction. Their theoretical and empirical results also support the superiority of LP relaxations in this setting.
\citet{martinsICML09} established conditions which guarantee algorithmic separability for LP relaxed training, and derived risk bounds for a learning algorithm which uses a combination of exact and relaxed inference.


Finally, recently \citet{globersonICML15} studied the performance of structured predictors for 2D grid graphs with binary labels from an information-theoretic point of view.
They proved lower bounds on the minimum achievable expected Hamming error in this setting, and proposed a polynomial-time algorithm that achieves this error.
Our work is different since we focus on LP relaxations as an approximation algorithm, we handle the most general form without making any assumptions on the model or error measure (except score decomposition), and we concentrate solely on the computational aspects while ignoring any accuracy concerns.

\section{Background}
In this section we review the formulation of the structured prediction problem, its LP relaxation, and the associated learning problem.
Consider a prediction task where the goal is to map a real-valued input vector $x$ to a discrete output vector $y=(y_1,\ldots,y_n)$.
A popular model class for this task is based on linear classifiers. In this setting prediction is performed via a linear discriminant rule: $y(x;w) = \argmax_{y'} w^\top\phi(x,y')$, where $\phi(x,y)\in\reals^d$ is a function mapping input-output pairs to feature vectors, and $w\in\reals^d$ is the corresponding weight vector.
Since the output space is often huge (exponential in $n$), it will generally be intractable to maximize over all possible outputs.

In many applications the score function has a particular structure. Specifically, we will assume that the score decomposes as a sum of simpler score functions: $w^\top\phi(x,y)=\sum_c w_c^\top\phi_c(x,y_c)$, where $y_c$ is an assignment to a (non-exclusive) subset of the variables $c$. 
For example, it is common to use such a decomposition that assigns scores to single and pairs of output variables corresponding to nodes and edges of a graph $G$: $w^\top\phi(x,y) = \sum_{i\in V(G)} w_i^\top\phi_i(x,y_i) + \sum_{ij\in E(G)} w_{ij}^\top\phi_{ij}(x,y_i,y_j)$.
Viewing this as a function of $y$, we can write the prediction problem as: $\max_y \sum_c \theta_c(y_c;x,w)$ (we will sometimes omit the dependence on $x$ and $w$ in the sequel).

Due to its combinatorial nature, the prediction problem is generally NP-hard. 
Fortunately, efficient approximations have been proposed.
Here we will be particularly interested in approximations based on LP relaxations.
We begin by formulating prediction as the following ILP:%
\footnote{For convenience we introduce singleton factors $\theta_i$, which can be set to $0$ if needed.}
{\small
\begin{align*}
\label{eq:lp_relax}
& \max_{\substack{\mu\in\localpoly \\ \mu\in\{0,1\}^q}} \sum_c \sum_{y_c} \mu_c(y_c) \theta_c(y_c) + \sum_i \sum_{y_i} \mu_i(y_i) \theta_i(y_i)
\quad = \theta^\top\mu \\
& \text{where } \localpoly = \left\{
\mu\geq 0 :
\hspace{-3pt}
\begin{array}{ll}
\sum_{y_{c\setminus i}} \mu_{c}(y_c) = \mu_{i}(y_i) & \hspace{-3pt}\forall c,i\in c, y_i\\
\sum_{y_i} \mu_{i}(y_i) = 1 & \hspace{-3pt}\forall i \\
\end{array}
\hspace{-3pt}
\right\}. \nonumber
\end{align*}
}%
Here, $\mu_c(y_c)$ is an indicator variable for a factor $c$ and local assignment $y_c$, and $q$ is the total number of factor assignments (dimension of $\mu$). The set $\localpoly$ is known as the local marginal polytope \cite{jordan}.
First, notice that there is a one-to-one correspondence between feasible $\mu$'s and assignments $y$'s, which is obtained by setting $\mu$ to indicators over local assignments ($y_c$ and $y_i$) consistent with $y$. 
Second, while solving ILPs is NP-hard in general, it is easy to obtain a tractable program by relaxing the integrality constraints $(\mu\in\{0,1\}^q)$, which may introduce fractional solutions to the LP.
This relaxation is the first level of the Sherali-Adams hierarchy \cite{sherali1990hierarchy}, which provides successively tighter LP relaxations of an ILP.
Notice that since the relaxed program is obtained by removing constraints, its optimal value upper bounds the ILP optimum.

In order to achieve high prediction accuracy, the parameters $w$ are learned from training data.
In this supervised learning setting, the model is fit to labeled examples $\{(x^{(m)},y^{(m)})\}_{m=1}^M$, where the goodness of fit is measured by a task-specific loss $\Delta(y(x^{(m)};w),y^{(m)})$.
In the \emph{structured SVM} (SSVM) framework \cite{Taskar03,Tsochantaridis04}, the empirical risk is upper bounded by a convex surrogate called the structured hinge loss, which yields the training objective:%
\footnote{For brevity, we omit the regularization term, however, all of our results below still hold with regularization.}
{\small
\begin{equation}
\label{eq:svm}
\min_w\sum_m \max_y \left[ w^\top\hspace{-3pt}\left(\phi(x^{(m)},y)-\phi(x^{(m)}, y^{(m)})\right) \hspace{-2pt} + \Delta(y,y^{(m)}) \right]~.
\end{equation}
}%
This is a convex function of $w$ and hence can be optimized in various ways. But, notice that the objective includes a maximization over outputs $y$ for each training example. This loss-augmented prediction task needs to be solved repeatedly during training (e.g., to evaluate subgradients), which makes training intractable in general.
Fortunately, as in prediction, LP relaxation can be applied to the structured loss \cite{Taskar03,Kulesza07}, 
which yields the relaxed training objective:
\begin{equation}
\label{eq:relaxed_svm}
\min_w\;\; \sum_m \max_{\mu\in\localpoly} \left[ \theta_m^\top (\mu - \mu_m) + \ell_m^\top \mu \right] ~,
\end{equation}
where $\theta_m \in \reals^q$ is a score vector in which each entry represents $w_c^\top \phi_c(x^{(m)},y_c)$ for some $c$ and $y_c$,
similarly $\ell_m \in \reals^q$ is a vector with entries%
\footnote{We assume that the task-loss $\Delta$ decomposes as the model score.}
$\Delta_c(y_c,y^{(m)}_c)$, 
and $\mu_m$ is the integral vector corresponding to $y^{(m)}$.

\section{Analysis}
\label{sec:analysis}
In this section we present our main results, proposing a theoretical justification for the observed tightness of LP relaxations used for inference in models learned by structured prediction, both on training and held-out data.
To this end, we make two complementary arguments:
in \secref{sec:train_integrality} we argue that optimizing the relaxed training objective of \eqref{eq:relaxed_svm} also has the effect of encouraging tightness of training instances;
in \secref{sec:test_integrality} we show that tightness generalizes from train to test data.

\subsection{Tightness at Training}
\label{sec:train_integrality}
We first show that the \emph{relaxed} training objective in \eqref{eq:relaxed_svm}, although designed to achieve high accuracy, also induces tightness of the LP relaxation.
In order to simplify notation we focus on a single training instance and drop the index $m$.
Denote the solutions to the relaxed and integer LPs as:
\begin{equation*}
\mu_L \in \argmax_{\mu\in\localpoly} \theta^\top \mu \qquad\qquad
\mu_I \in \argmax_{\substack{\mu\in\localpoly \\ \mu\in\{0,1\}^q}} \theta^\top \mu
\end{equation*}
Also, let $\mu_T$ be the integral vector corresponding to the ground-truth output $y^{(m)}$.
Now consider the following decomposition:
\begin{equation}
\label{eq:hinge_decomposition}
\underset{\text{\color{blue}{relaxed-hinge}}}{\theta^\top(\mu_L - \mu_T)} =
\underset{\text{\color{blue}{integrality gap}}}{\theta^\top(\mu_L - \mu_I)} + \underset{\text{\color{blue}{exact-hinge}}}{\theta^\top(\mu_I - \mu_T)}
\end{equation}
This equality states that the difference in scores between the relaxed optimum and ground-truth (\emph{relaxed-hinge}) can be written as a sum of the \emph{integrality gap} and the difference in scores between the exact optimum and the ground-truth (\emph{exact-hinge}) (notice that all terms are non-negative).
This simple decomposition has several interesting implications.

First, we can immediately derive the following bound on the integrality gap:
{\small
\begin{align}
\theta^\top(\mu_L - \mu_I)
=&~ \theta^\top(\mu_L - \mu_T) - \theta^\top(\mu_I - \mu_T) \label{eq:train_gap_bound_exact_form} \\
\le&~ \theta^\top(\mu_L - \mu_T) \label{eq:train_gap_bound_remove_exact_hinge} \\
\le&~ \theta^\top(\mu_L - \mu_T) + \ell^\top\mu_L \label{eq:train_gap_bound_add_loss} \\
\le& \max_{\mu\in\localpoly} \left( \theta^\top(\mu - \mu_T) + \ell^\top\mu \right), 
\label{eq:train_gap_loss_bound}
\end{align}
}%
where \eqref{eq:train_gap_loss_bound} is precisely the relaxed training objective from \eqref{eq:relaxed_svm}.
Therefore, optimizing the approximate training objective of \eqref{eq:relaxed_svm} \emph{minimizes an upper bound on the integrality gap}.
Hence, driving down the approximate objective also reduces the integrality gap of training instances.
One case where the integrality gap becomes zero is when the data is algorithmically separable. 
In this case the relaxed-hinge term vanishes (the exact-hinge must also vanish), and integrality is assured.

However, the bound above might sometimes be loose.
Indeed, to get the bound we have discarded the exact-hinge term (\eqref{eq:train_gap_bound_remove_exact_hinge}), added the task-loss (\eqref{eq:train_gap_bound_add_loss}), and maximized the loss-augmented objective (\eqref{eq:train_gap_loss_bound}).
At the same time, \eqref{eq:train_gap_bound_exact_form} provides a precise characterization of the integrality gap.
Specifically, the gap is determined by the difference between the relaxed-hinge and the exact-hinge terms.
This implies that even when the relaxed-hinge is not zero, a small integrality gap can still be obtained if the exact-hinge is also large.
In fact, the \emph{only way} to get a large integrality gap is by setting the exact-hinge much smaller than the relaxed-hinge.
But when can this happen?

A key point is that the relaxed and exact hinge terms are upper bounded by the relaxed and exact \emph{training objectives}, respectively (the latter additionally depend on the task loss $\Delta$).
Therefore, minimizing the training objective will also reduce the corresponding hinge term (see also \secref{sec:experiments}).
Using this insight, we observe that relaxed training reduces the relaxed-hinge term without directly reducing the exact-hinge term, and thereby induces a small integrality gap.
On the other hand, this also suggests that \emph{exact training may actually increase the integrality gap}, since it reduces the exact-hinge without also reducing directly the relaxed-hinge term.
This finding is consistent with previous empirical evidence. Specifically, \citet[][Table 2]{martinsICML09} showed that on a dependency parsing problem, training with the relaxed objective achieved $92.88\%$ integral solutions, while exact training achieved only $83.47\%$ integral solutions.
An even stronger effect was observed by \citet[][Table 3]{Finley08} for multi-label classification, where relaxed training resulted in $99.57\%$ integral instances, with exact training attaining only $17.7\%$ (`Yeast' dataset).

In \secref{sec:experiments} we provide further empirical support for our explanation, however, we next also show its possible limitations by providing a counter-example.
The counter-example demonstrates that despite training with a relaxed
objective, the exact-hinge can in some cases actually be {\em smaller} than the
relaxed-hinge, leading to a loose relaxation.
Although this illustrates the limitations of the explanation above, we point out that the corresponding learning task is far from natural; %
we believe it is unlikely to arise in real-world applications.

Specifically, we construct a learning scenario where relaxed training obtains zero exact-hinge and non-zero relaxed-hinge, so the relaxation is not tight.
%
Consider a model where $x\in\reals^3$, $y\in\{0,1\}^3$, and the prediction is given by:
\vspace{-2mm}
\begin{align*}
& y(x;w) = \argmax_y \Big( x_1 y_1 + x_2 y_2 + x_3 y_3 \\
&\quad + w \left[ \mathbbm{1}\{y_1\ne y_2\} + \mathbbm{1}\{y_1\ne y_3\} + \mathbbm{1}\{y_2\ne y_3\} \right] \Big).
\end{align*}
\vspace{-2mm}
The corresponding LP relaxation is then:
{{\small \begin{align*}
& \max_{\mu\in\localpoly} \Big( x_1\mu_1(1) + x_2\mu_2(1) + x_3\mu_3(1) + w [ \mu_{12}(01)+\mu_{12}(10) \\
&\quad + \mu_{13}(01)+\mu_{13}(10) + \mu_{23}(01)+\mu_{23}(10)] \Big).
\end{align*}}}%
Next, we construct a trainset where the first instance is: $x^{(1)}=(2,2,2), y^{(1)}=(1,1,0)$, and the second is: $x^{(2)}=(0,0,0), y^{(2)}=(1,1,0)$.
It can be verified that $w=1$ minimizes the relaxed objective (\eqref{eq:relaxed_svm}).
However, with this weight vector the relaxed-hinge for the second instance is equal to 1, while the exact-hinge for both instances is 0 (the data is separable w.r.t.~$w=1$).
Consequently, there is an integrality gap of 1 for the second
instance, and the relaxation is loose (the first instance is actually
tight).

Finally, note that our derivation above (\eqref{eq:train_gap_bound_exact_form}) holds for \emph{any integral $\mu$}, and not just the ground-truth $\mu_T$.
In other words, the only property of $\mu_T$ we are using here is its integrality.
Indeed, in \secref{sec:experiments} we verify empirically that training a model using \emph{random labels} still attains the same level of tightness as training with the ground-truth labels.
On the other hand, accuracy drops dramatically, as expected.
This analysis suggests that \emph{tightness is not related to accuracy} of the predictor.
\citet{Finley08} explained tightness of LP relaxations by noting that fractional solutions always incur a loss during training.
Our analysis suggests an alternative explanation, emphasizing the difference in scores (\eqref{eq:train_gap_bound_exact_form}) rather than the loss, and decoupling tightness from accuracy.

\subsection{Generalization of Tightness}
\label{sec:test_integrality}
Our argument in \secref{sec:train_integrality} concerns only the tightness of train instances.
However, the empirical evidence discussed above pertains to test data.
To bridge this gap, in this section we show that train tightness implies test tightness. We do so by proving a generalization bound for tightness based on Rademacher complexity.

We first define a loss function which measures the lack of integrality (or, fractionality) for a given instance.
To this end, we consider the discrete set of
\emph{vertices} of the local
polytope $\localpoly$ (excluding its convex hull), denoting by
${\cal{M}}^I$ and ${\cal{M}}^F$ the sets of fully-integral and
non-integral (i.e., fractional) vertices, respectively (so
${\cal{M}}^I \cap {\cal{M}}^F = \emptyset$, and ${\cal{M}}^I \cup
{\cal{M}}^F$ consists of all vertices of $\localpoly$).
Considering vertices is without loss of generality, since 
  linear programs always have a vertex that is optimal. 
Next, let $\theta_x \in \reals^q$ be the mapping from weights $w$ and inputs $x$ to scores (as used in \eqref{eq:relaxed_svm}), and
let $I^*(\theta)=\max_{\mu\in{\cal{M}}^I}\theta^\top\mu$ and $F^*(\theta)=\max_{\mu\in{\cal{M}}^F}\theta^\top\mu$ be the best integral and fractional scores attainable, respectively.
By convention, we set $F^*(\theta)=-\infty$ whenever ${\cal{M}}^F = \emptyset$.
The fractionality of $\theta$ can be measured by the quantity
$D(\theta) = F^*(\theta) - I^*(\theta)$.
If this quantity is large then the LP has a fractional solution with a much better score than any integral solution.
We can now define the loss:
\vspace{-2mm}
\begin{equation}
{\cal{L}}(\theta) = \begin{cases}
1 & D(\theta)>0 \\
0 & \text{otherwise}
\end{cases}~.
\vspace{-1mm}
\end{equation}
That is, the loss equals $1$ if and only if the optimal fractional solution has a (strictly) higher score than the optimal integral solution.%
\footnote{Notice that the loss will be $0$ whenever the non-integral and integral optima are equal, but this is fine for our purpose, since we consider the relaxation to be tight in this case.}
Notice that this loss ignores the ground-truth $y$, as expected.
In addition, we define a ramp loss parameterized by $\gamma>0$ which upper bounds the fractionality loss:
\vspace{-3mm}
\begin{equation}
\label{eq:ramp_loss}
\varphi_\gamma(\theta) = \begin{cases}
0 & D(\theta) \leq -\gamma \\
1+ D(\theta)/\gamma & -\gamma < D(\theta) \leq 0 \\
1 & D(\theta) > 0
\end{cases}~,
\end{equation}
For this loss to be zero, the best integral solution has to be better than the best fractional solution by at least $\gamma$, which is a stronger requirement than mere tightness.
In \secref{sec:gamma_tight} we give examples of models that are guaranteed to satisfy this stronger requirement, and in \secref{sec:experiments} we also show this often happens in practice.
We point out that $\varphi_\gamma(\theta)$ is generally hard to compute,
as is ${\cal{L}}(\theta)$ (due to the discrete optimization involved in computing $I^*(\theta)$ and $F^*(\theta)$).
However, here we are only interested in proving that tightness is a generalizing property, so we will not worry about computational efficiency for now. 
We are now ready to state the main theorem of this section. 

\begin{theorem}
\label{thm:integrality_generalization_bound}
Let inputs be independently selected according to a probability measure $P(X)$, and let $\Theta$ be the class of all scoring functions $\theta_X$ with $\|w\|_2 \leq B$.
Let $\|\phi(x,y_c)\|_2 \leq \hat R$ for all $x$, $c$, $y_c$, and $q$ is the total number of factor assignments (dimension of $\mu$).
Then for any number of samples $M$ and any $0 < \delta < 1$, with probability at least $1-\delta$, every $\theta_X\in\Theta$ satisfies:
{\small \begin{equation}
\label{eq:generalization_bound}
\expect_P [{\cal{L}}(\theta_X)] \leq \hat\expect_M [\varphi_\gamma(\theta_X)] + O\left(\frac{q^{1.5} B \hat R}{\gamma\sqrt{M}}\right) + \sqrt{\frac{8\ln(2/\delta)}{M}}
\end{equation}}%
\end{theorem}
where $\hat\expect_M$ is the empirical expectation.
\begin{proof}
Our proof 
relies on the following general result from \citet{bartlett2002rademacher}.
\begin{theorem}[\citet{bartlett2002rademacher}, Theorem 8]
\label{thm:gen_bound}
Consider a loss function ${\cal{L}}:{\cal{Y}}\times\Theta \mapsto [0,1]$ and a dominating function $\varphi:{\cal{Y}}\times\Theta \mapsto [0,1]$ (i.e., ${\cal{L}}(y,\theta) \leq \varphi(y,\theta)$ for all $y,\theta$). Let ${\cal{F}}$ be a class of functions mapping ${\cal{X}}$ to $\Theta$, and let $\{(x^{(m)},y^{(m)})\}_{m=1}^M$ be independently selected according to a probability measure %
$P(x,y)$.
Then for any number of samples $M$ and any $0 < \delta < 1$, with probability at least $1-\delta$, every $f\in {\cal{F}}$ satisfies:
{\small \[
\expect [{\cal{L}}(y,f(x))] \leq \hat\expect_M [\varphi(y,f(x))] + {\cal{R}}_M(\tilde\varphi \circ f) + \sqrt{\frac{8\ln(2/\delta)}{M}}~, 
\]}%
where $\hat\expect_M$ is the empirical expectation, $\tilde\varphi \circ f = \{(x,y)\mapsto \varphi(y,f(x)) - \varphi(y,0) : f\in {\cal{F}}\}$, and ${\cal{R}}_M({\cal{F}})$ is the Rademacher complexity of the class ${\cal{F}}$.
\end{theorem}

To use this result, we define $\Theta=\reals^q$, $f(x) = \theta_x$, and ${\cal{F}}$ to be the class of all such functions satisfying $\|w\|_2 \leq B$ and $\|\phi(x,y_c)\|_2 \leq \hat R$.
In order to obtain a meaningful bound, we would like to bound the Rademacher term ${\cal{R}}_M(\tilde\varphi \circ f)$.
Theorem 12 in \citet{bartlett2002rademacher} states that if $\tilde\varphi$ is Lipschitz with constant $L$ and satisfies $\tilde\varphi(0) = 0$, then ${\cal{R}}_M(\tilde\varphi \circ f) \leq 2L{\cal{R}}_M({\cal{F}})$.
In addition, \citet{weiss10cascades} show that ${\cal{R}}_M({\cal{F}}) = O(\frac{q B \hat R}{\sqrt{M}})$.
Therefore, it remains to compute the Lipschitz constant of $\tilde\varphi$, which is equal to the Lipschitz constant of $\varphi$. 
For this purpose, we will bound the Lipschitz constant of $D(\theta)$, and then use $L(\varphi_\gamma(\theta)) \le L(D(\theta))/\gamma$ (from \eqref{eq:ramp_loss}).\\
Let $\mu_I\in\argmax_{\mu\in{\cal{M}}^I}\theta^\top\mu$ and $\mu_F\in\argmax_{\mu\in{\cal{M}}^F}\theta^\top\mu$, then:
{\small \begin{align*}
& D(\theta^1) - D(\theta^2) \\
=&~ (\mu_F^1 - \mu_I^1) \cdot \theta^1 - ( \mu_F^2 - \mu_I^2) \cdot\theta^2 \\
=&~ ( \mu_F^1 \cdot \theta^1 - \mu_F^2\cdot\theta^2 ) + ( \mu_I^2\cdot\theta^2 - \mu_I^1\cdot\theta^1 ) \\
=&~ (\mu_F^1 \cdot \theta^1 - \mu_F^2\cdot\theta^2) + (\mu_F^1\cdot\theta^2 - \mu_F^1\cdot\theta^2) \\
&~ + (\mu_I^2\cdot\theta^2 - \mu_I^1\cdot\theta^1) + (\mu_I^2\cdot\theta^1 - \mu_I^2\cdot\theta^1) \\
=&~ \mu_F^1 \cdot (\theta^1 - \theta^2) + (\mu_F^1 - \mu_F^2)\cdot\theta^2 \\
&~ + \mu_I^2\cdot (\theta^2 - \theta^1) + (\mu_I^2 - \mu_I^1) \cdot\theta^1 \\
\leq&~ ( \mu_F^1 - \mu_I^2) \cdot (\theta^1 - \theta^2) \qquad\quad\; [\text{optimality of } \mu_F^2 \text{ and } \mu_I^1] \\
\leq&~ \| \mu_F^1 - \mu_I^2 \|_2 \| \theta^1 - \theta^2 \|_2   \qquad\qquad\qquad\; [\text{Cauchy-Schwarz}] \\
\leq&~ \sqrt{q} \| \theta^1 - \theta^2 \|_2
\end{align*} }%
Therefore, $L=\sqrt{q}/\gamma$.

Combining everything together, and dropping the spurious dependence on $y$, we obtain the bound in \eqref{eq:generalization_bound}.
Finally, we point out that when
using an $L_2$ regularizer at training,
we can actually drop the assumption $\|w\|_2\le B$ and instead use a bound on the norm of the optimal solution (as in the analysis of \citet{shalev2011pegasos}).
%
%
\end{proof}

\thmref{thm:integrality_generalization_bound}
shows that if we observe high integrality (equivalently, low
fractionality) on a finite sample of training data, then it is likely
that integrality of test data will not be much lower, provided
sufficient number of samples.
%

Our result actually applies more generally to any two disjoint sets of vertices, and is not limited to ${\cal{M}}^I$ and ${\cal{M}}^F$.
For example, we can replace ${\cal{M}}^I$ by the set of vertices with at most 10\% fractional values, and ${\cal{M}}^F$ by the rest of the vertices of the local polytope.
This gives a different meaning to the loss $D(\theta)$, and the rest of our analysis holds unchanged.
Consequently, our generalization result implies that it is likely to observe a similar portion of instances with at most 10\% fractional values at test time as we did at training.

\subsubsection{$\gamma$-tight relaxations}
\label{sec:gamma_tight}
In this section we study the stronger notion of tightness required by our surrogate fractionality loss (\eqref{eq:ramp_loss}), and show examples of models that satisfy it.
We use the following definition.
\vspace{-2mm}
\begin{definition} An LP relaxation is called \emph{$\gamma$-tight} if $I^*(\theta) \ge F^*(\theta) + \gamma$ (so 
$\varphi_\gamma(\theta)=0$). That is, the best integral value is larger than the best non-integral value by at least $\gamma$.%
\footnote{Notice that scaling up $\theta$ will also increase $\gamma$, but our bound in \eqref{eq:generalization_bound} also grows with the norm of $\theta$ (via $B\hat R$). Therefore, we assume here that $\|\theta\|_2$ is bounded.}
\end{definition}
\vspace{-2mm}
We focus on binary pairwise models and show two cases where the model is guaranteed to be $\gamma$-tight.
Proofs are provided in \appref{app:gamma_tight}.
Our first example involves \emph{balanced} models, which are binary pairwise models that have supermodular scores, or can be made supermodular by ``flipping'' a subset of the variables (for more details, see \appref{app:gamma_tight}).
\begin{proposition}
\label{prop:balanced}
A balanced model with a unique optimum is $(\alpha/2)$-tight, where $\alpha$ is the difference between the best and second-best (integral) solutions.
\end{proposition}

This result is of particular interest when learning structured
predictors where the edge scores depend on the input.
Whereas one could learn supermodular models by enforcing linear inequalities,
we know of no tractable means of restricting the model to be balanced. Instead, one could learn over the full
space of models using LP relaxation. If the learned models are
balanced on the training data, Prop.~\ref{prop:balanced} together with 
Theorem~\ref{thm:integrality_generalization_bound} tell us that the
pairwise LP relaxation is likely to be tight on test data as well.

Our second example regards models with singleton scores that are much stronger than the pairwise scores.
Consider a binary pairwise model%
\footnote{This case easily generalizes to non-binary variables.}
in minimal representation, where $\bar\theta_i$ are node scores and $\bar\theta_{ij}$ are edge scores in this representation (see \appref{app:gamma_tight} for full details).
Further, for each variable $i$, define the set of neighbors with \emph{attractive} edges $N_i^+=\{j\in N_i | \bar\theta_{ij}>0\}$, and the set of neighbors with \emph{repulsive} edges $N_i^-=\{j\in N_i | \bar\theta_{ij}<0\}$.
\begin{proposition}
\label{prop:strong_singles}
If all variables satisfy the condition:
\vspace{-5pt}
\[
\bar\theta_i \ge -\sum_{j\in N_i^-} \bar\theta_{ij} + \beta
,\quad\text{or } \quad
\bar\theta_i \le -\sum_{j\in N_i^+} \bar\theta_{ij} - \beta
\vspace{-5pt}
\]
for some $\beta>0$, then the model is $(\beta/2)$-tight.
\end{proposition}

\vspace{-5pt}
Finally, we point out that in both of the examples above, the conditions can be verified efficiently and if they hold, the value of $\gamma$ can be computed efficiently.

\section{Experiments}
\label{sec:experiments}

\begin{figure*}
 \begin{center}
  \begin{tabular}{c c c}
  \hspace{-30px}
  \includegraphics[height=1.7in]{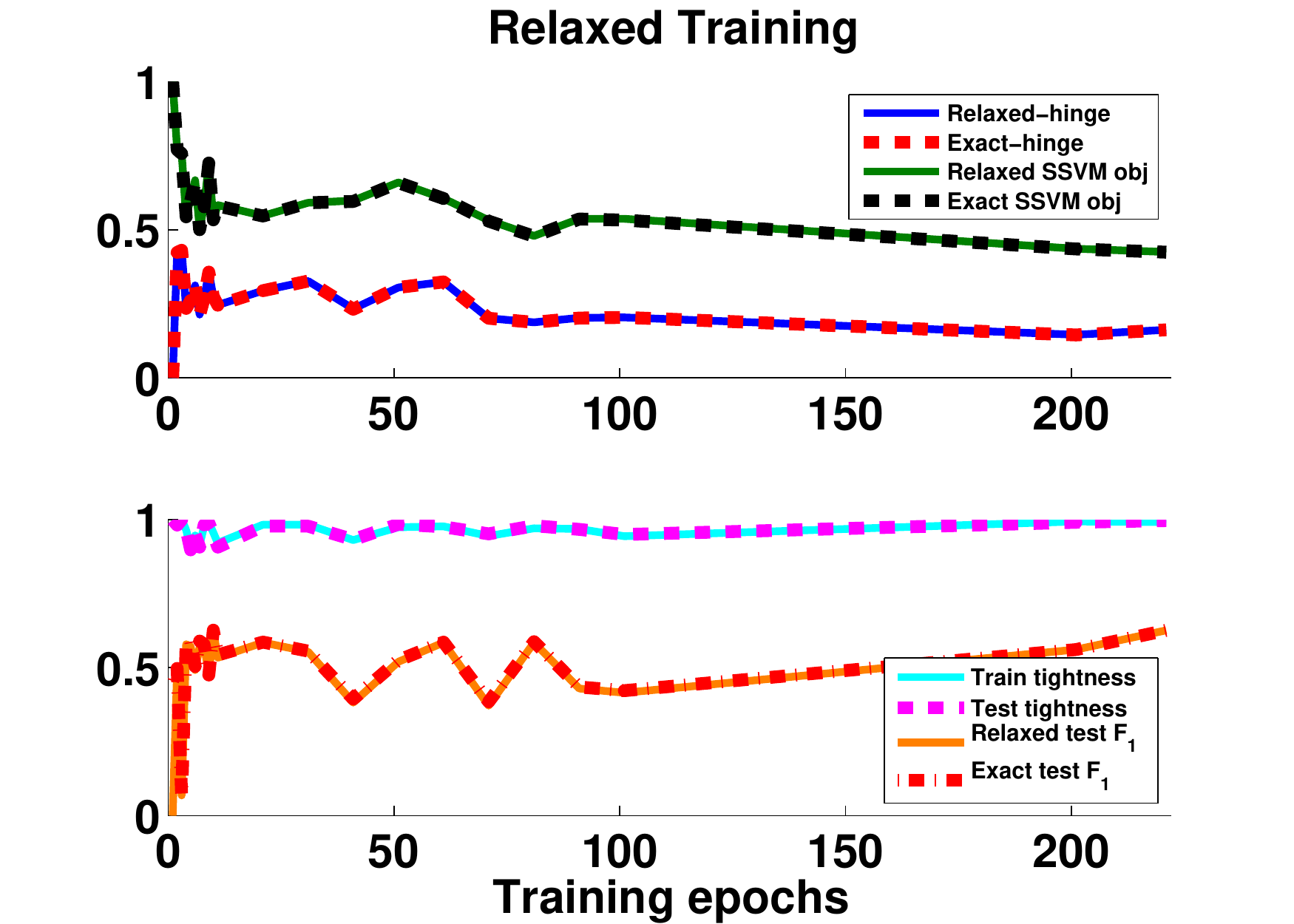}
  &
  \hspace{-25px}
  \includegraphics[height=1.7in]{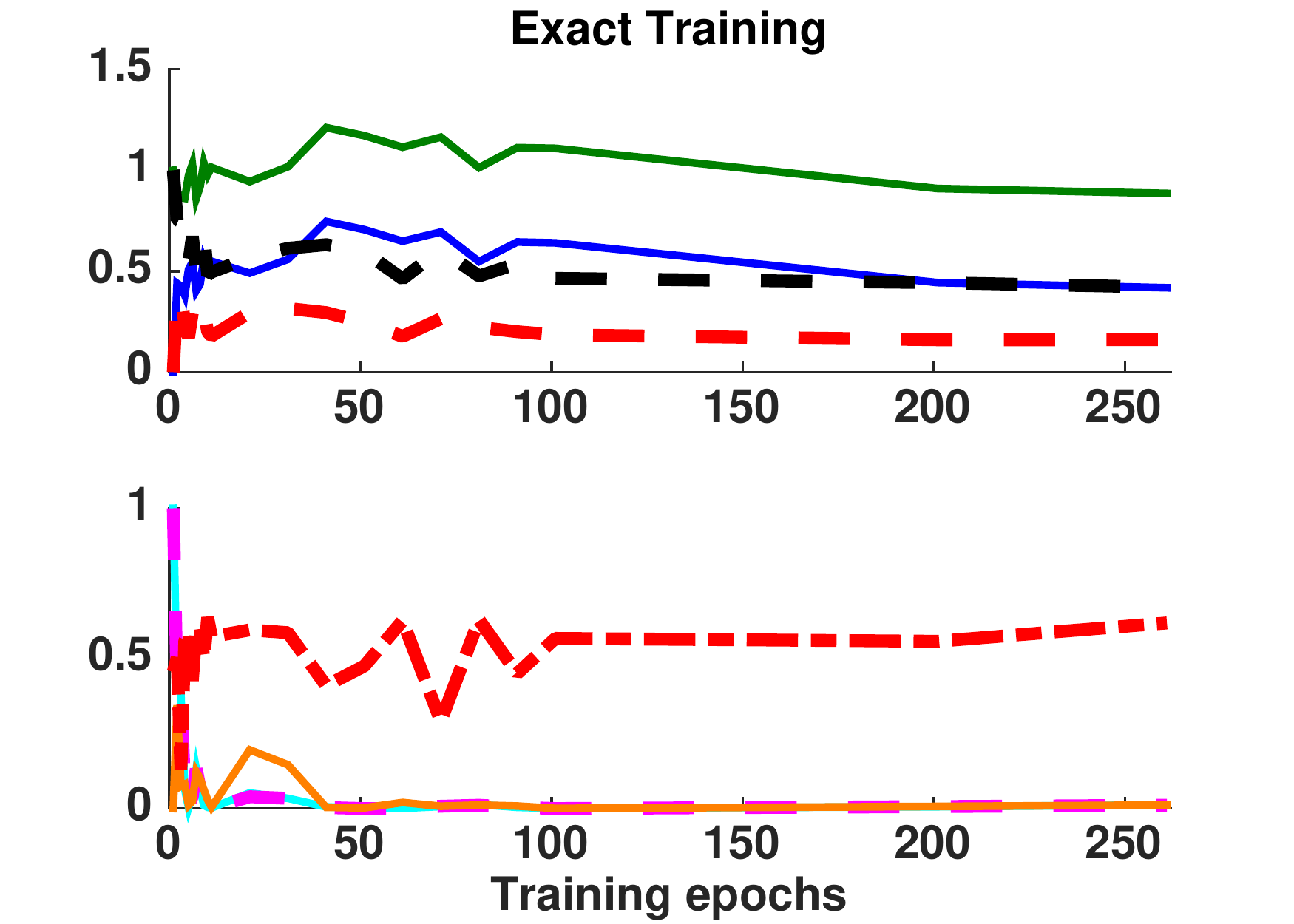}
  &
  \hspace{-13px}
  \includegraphics[height=1.7in]{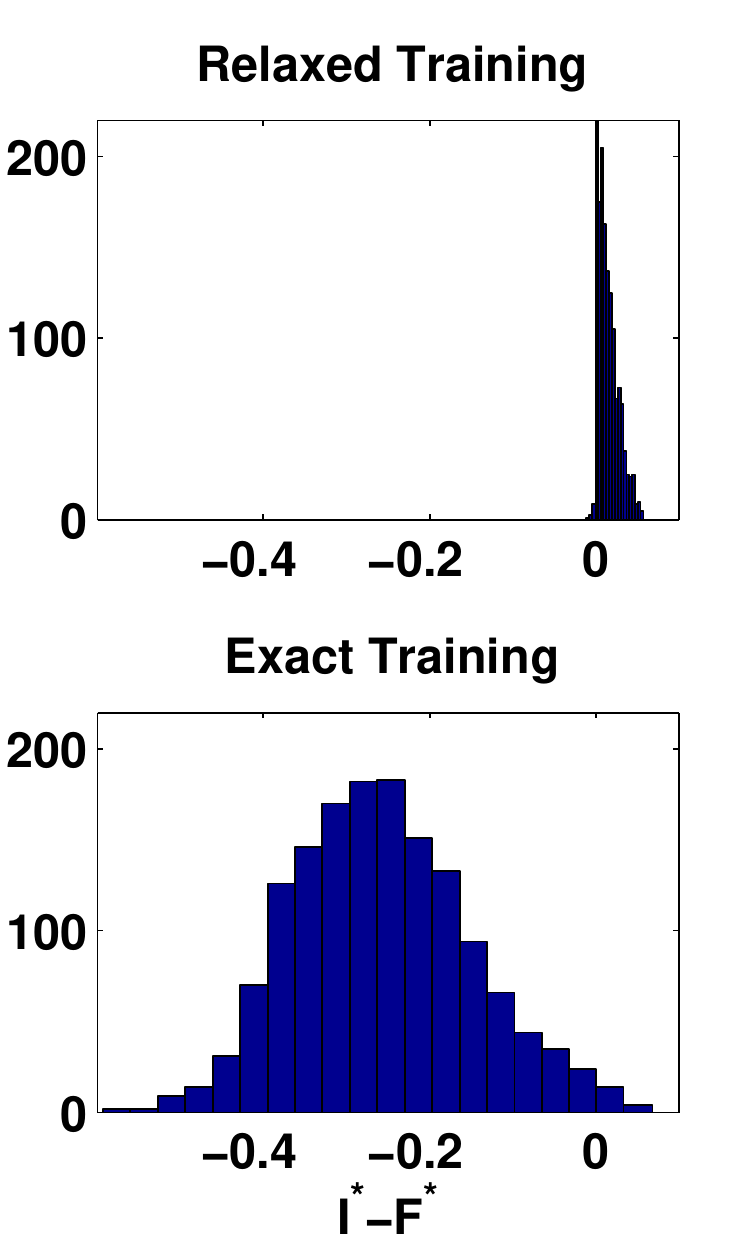}
  \end{tabular}
 \end{center}
\vspace{-4mm}
 \caption{{Training with the `Yeast' dataset. Various quantities of interest are shown as a function of training iterations. (Left) Training with LP relaxation. (Middle) Training with ILP. (Right) Integrality margin (bin widths are scaled differently).}}
  \label{fig:yeast}
\vspace{-0.3cm}
\end{figure*}

In this section we present some numerical results to support our theoretical analysis.
We run experiments for both a multi-label classification task and an image segmentation task.
For training we have implemented the block-coordinate Frank-Wolfe algorithm for structured SVM \cite{lacoste-julien13}, using GLPK as the LP solver.%
\footnote{\url{http://www.gnu.org/software/glpk}}
In all of our experiments we use a standard $L_2$ regularizer, 
chosen via cross-validation.

\paragraph{Multi-label classification}
For multi-label classification we adopt the experimental setting of \citet{Finley08}.
In this setting labels are represented by binary variables, the model consists of singleton and pairwise factors forming a fully connected graph over the labels,
and the task loss is the normalized Hamming distance. 

\figref{fig:yeast} shows relaxed and exact training iterations for the `Yeast' dataset (14 labels).
We plot the relaxed and exact hinge terms (\eqref{eq:hinge_decomposition}), the exact and relaxed SSVM training objectives%
\footnote{The displayed objective values are averaged over train instances and exclude regularization.}
(\eqref{eq:svm} and \eqref{eq:relaxed_svm}, respectively), 
fraction of train and test instances having integral solutions, as well as test accuracy (measured by $F_1$ score).
Whenever a fractional solution was found with relaxed inference, a simple rounding scheme was applied to obtain a valid prediction.
First, we note that the relaxed-hinge values are nicely correlated with the relaxed training objective, and likewise the exact-hinge is correlated with the exact objective (left and middle, top). 
Second, observe that with relaxed training, the relaxed-hinge and the exact-hinge are very close (left, top), so the integrality gap, given by their difference, remains small (almost $0$ here).
On the other hand, with exact training the exact-hinge is reduced much more than the relaxed-hinge, which results in a large integrality gap (middle, top).
Indeed, we can see that the percentage of integral solutions is almost $100\%$ for relaxed training (left, bottom), and close to $0\%$ with exact training (middle, bottom).
To get a better understanding, we show a histogram of the difference between the optimal integral and fractional values, i.e., the integrality margin ($I^*(\theta)-F^*(\theta)$), under the final learned model for all training instances (right). It can be seen that with relaxed training this margin is positive (although small), while exact training results in larger negative values.
Third, we notice that train and test integrality levels are very close to each other, almost indistinguishable (left and middle, bottom), which provides some empirical support to our generalization result from \secref{sec:test_integrality}.

\begin{figure*}
 \begin{center}
  \begin{tabular}{c c c}
  \hspace{-30px}
  \includegraphics[height=1.7in]{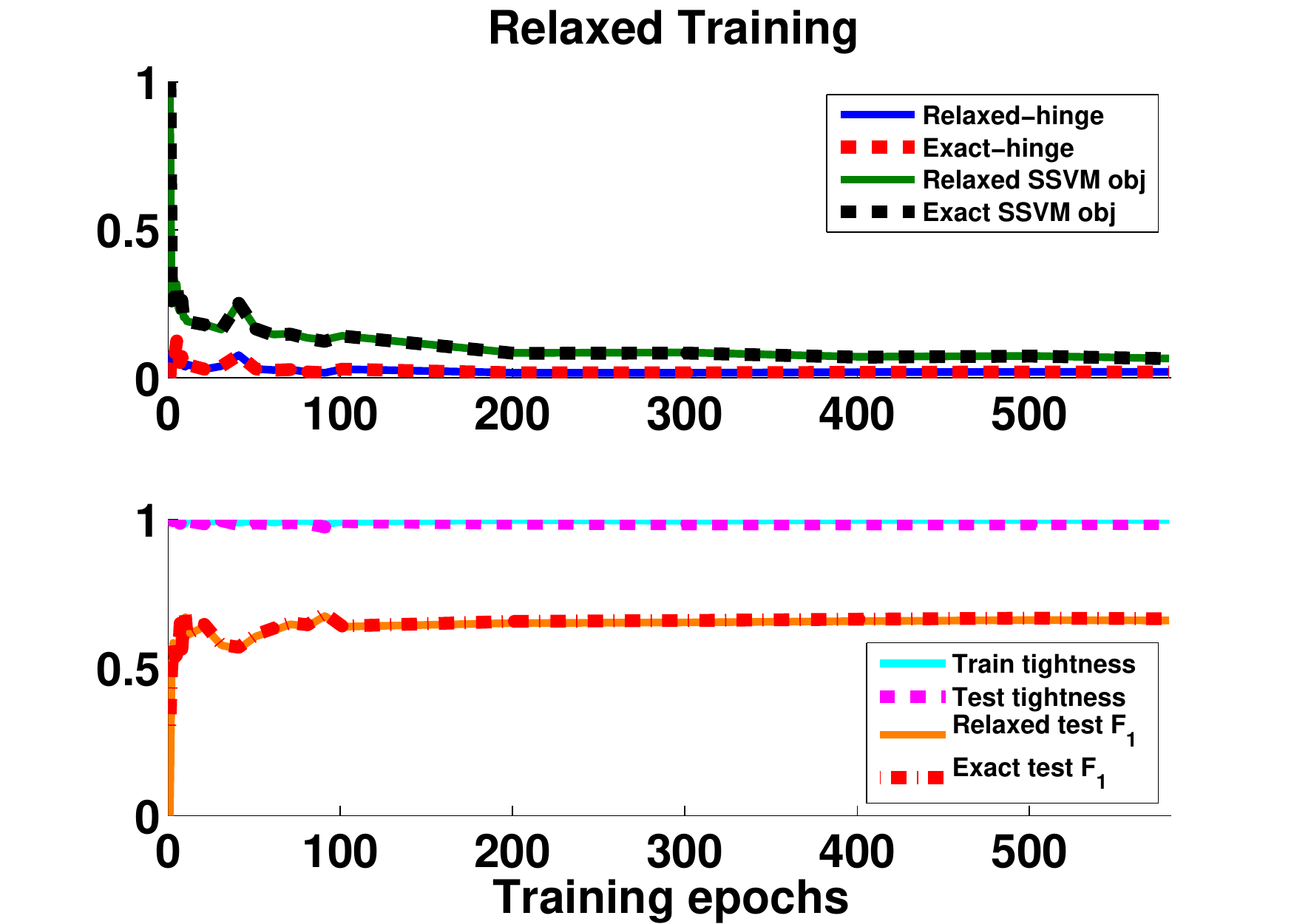}
  &
  \hspace{-25px}
  \includegraphics[height=1.7in]{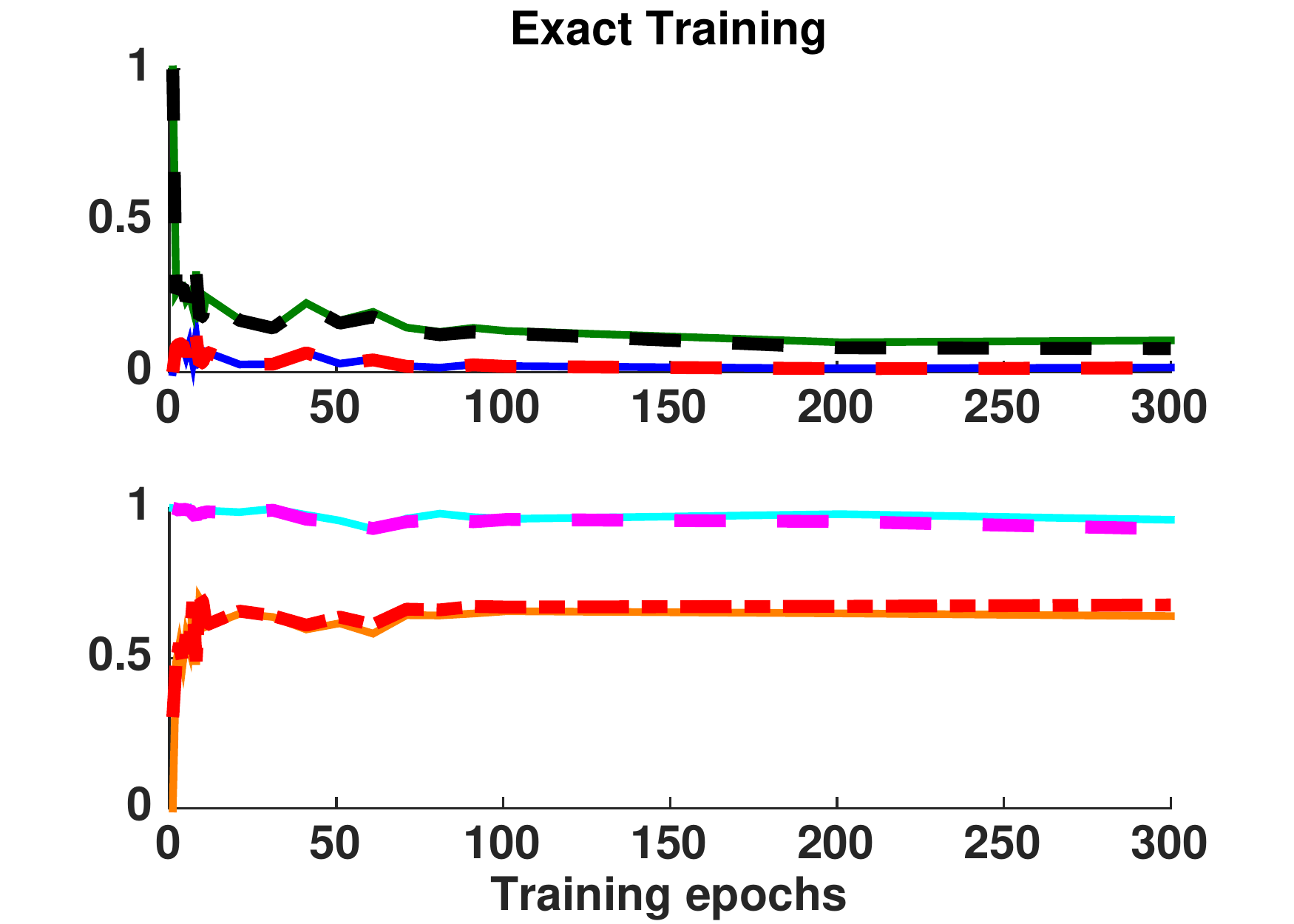}
  &
  \hspace{-10px}
  \includegraphics[height=1.7in]{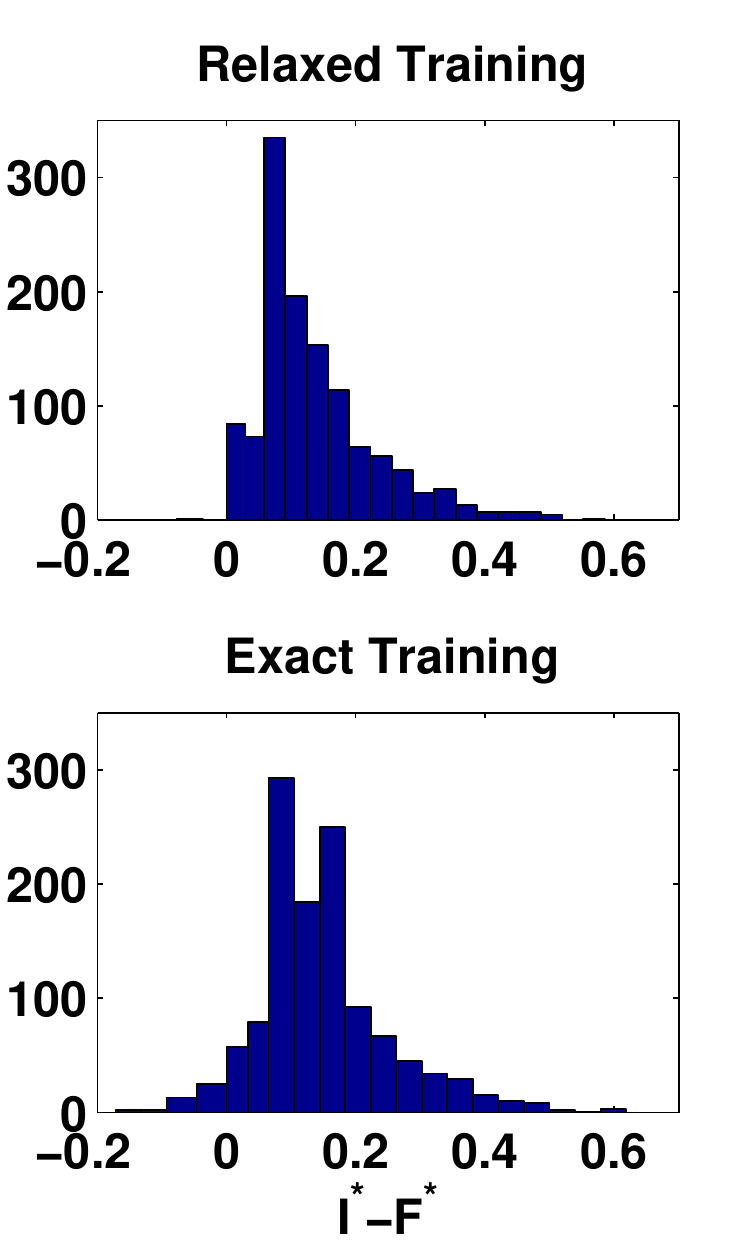}
  \end{tabular}
 \end{center}
\vspace{-4mm}
 \caption{{Training with the `Scene' dataset. Various quantities of interest are shown as a function of training iterations. (Left) Training with LP relaxation. (Middle) Training with ILP. (Right) Integrality margin.}}
  \label{fig:scene}
\end{figure*}


We next train a model using random labels (with similar label counts as the true data).
In this setting the learned model obtains $100\%$ tight training instances (not shown), which supports our claim that any integral solution can be used in place of the ground-truth, and that accuracy is not important for tightness.
Finally, in order to verify that tightness is not coincidental, we tested the tightness of the relaxation induced by a random weight vector $w$.
We found that random models are never tight (in 20 trials), which shows that tightness of the relaxation does not come by chance.

We now proceed to perform experiments on the `Scene' dataset (6 labels).
The results, in \figref{fig:scene}, are quite similar to the `Yeast' results, except for the behavior of exact training (middle) and the integrality margin (right).
Specifically, we observe that in this case the relaxed-hinge and exact-hinge are close in value (middle, top), as for relaxed training (left, top).
As a consequence, the integrality gap is very small and the relaxation is tight for almost all train (and test) instances.
These results show that sometimes optimizing the exact objective can reduce the relaxed objective (and relaxed-hinge) as well.
Further, in this setting we observe a larger integrality margin (right), which means that the integral optimum is strictly better than the fractional one. 

We conjecture that the LP instances are easy in this case due to the dominance of the singleton scores.%
\footnote{With ILP training, the condition in Prop.~\ref{prop:strong_singles} is satisfied for 65\% of all variables, although only 1\% of the training instances satisfy it for all their variables.}
Specifically, the features provide a strong signal which allows label assignment to be decided mostly based on the local score, with little influence coming from the pairwise terms.
To test this conjecture we repeat the experiment while injecting Gaussian noise into the input features, forcing the model to rely more on the pairwise interactions.
We find that with the noisy singleton scores the results are indeed similar to the `Yeast' dataset, where a large integrality gap is observed and fewer instances are tight (see \appref{app:scene_noisy} in the supplement). 
%
%
%

\begin{figure*}[t]
 \begin{center}
  \begin{tabular}{c c}
  \hspace{-25px}
  \includegraphics[width=3in]{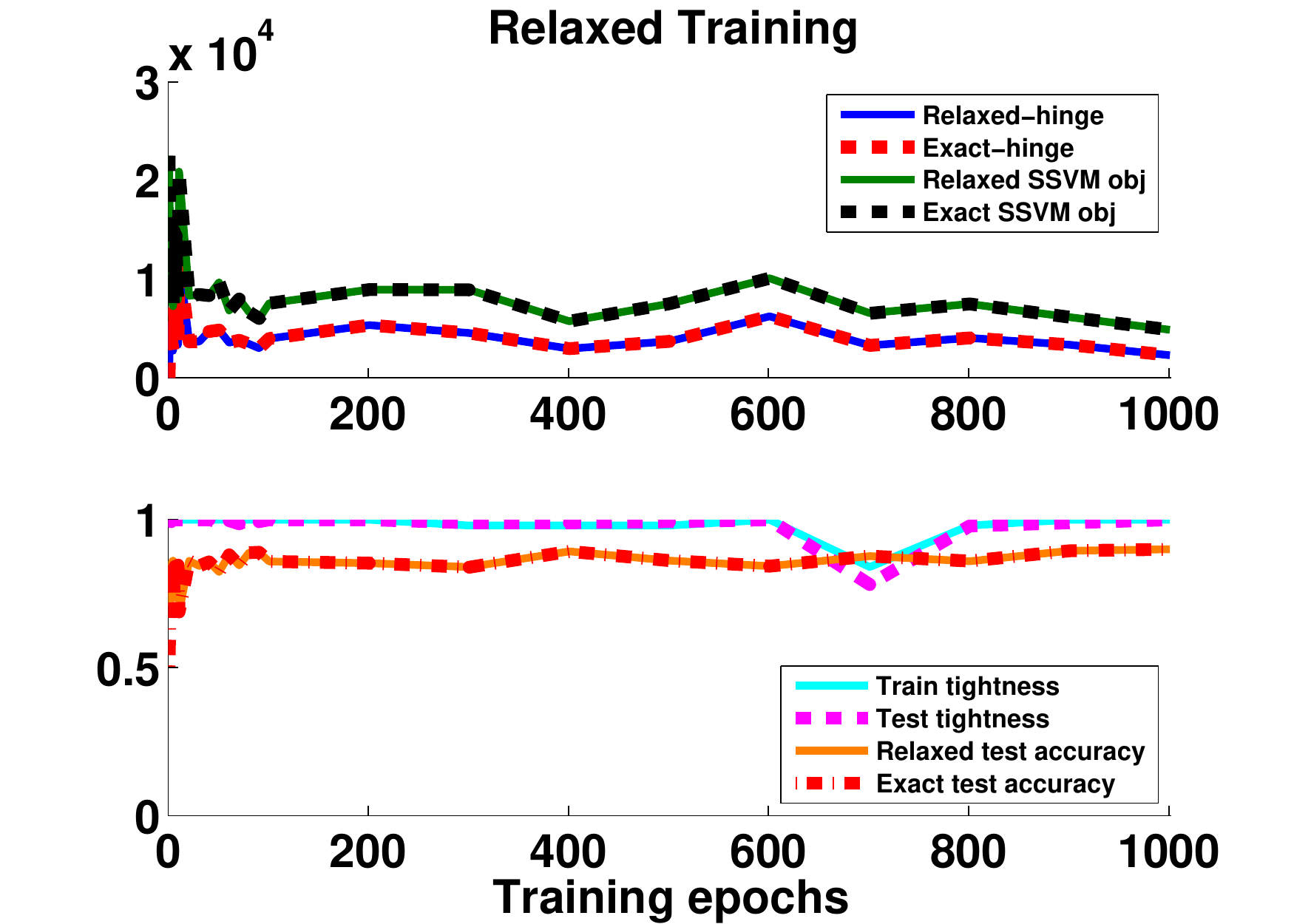}
  &
  \hspace{-30px}
  \includegraphics[width=3in]{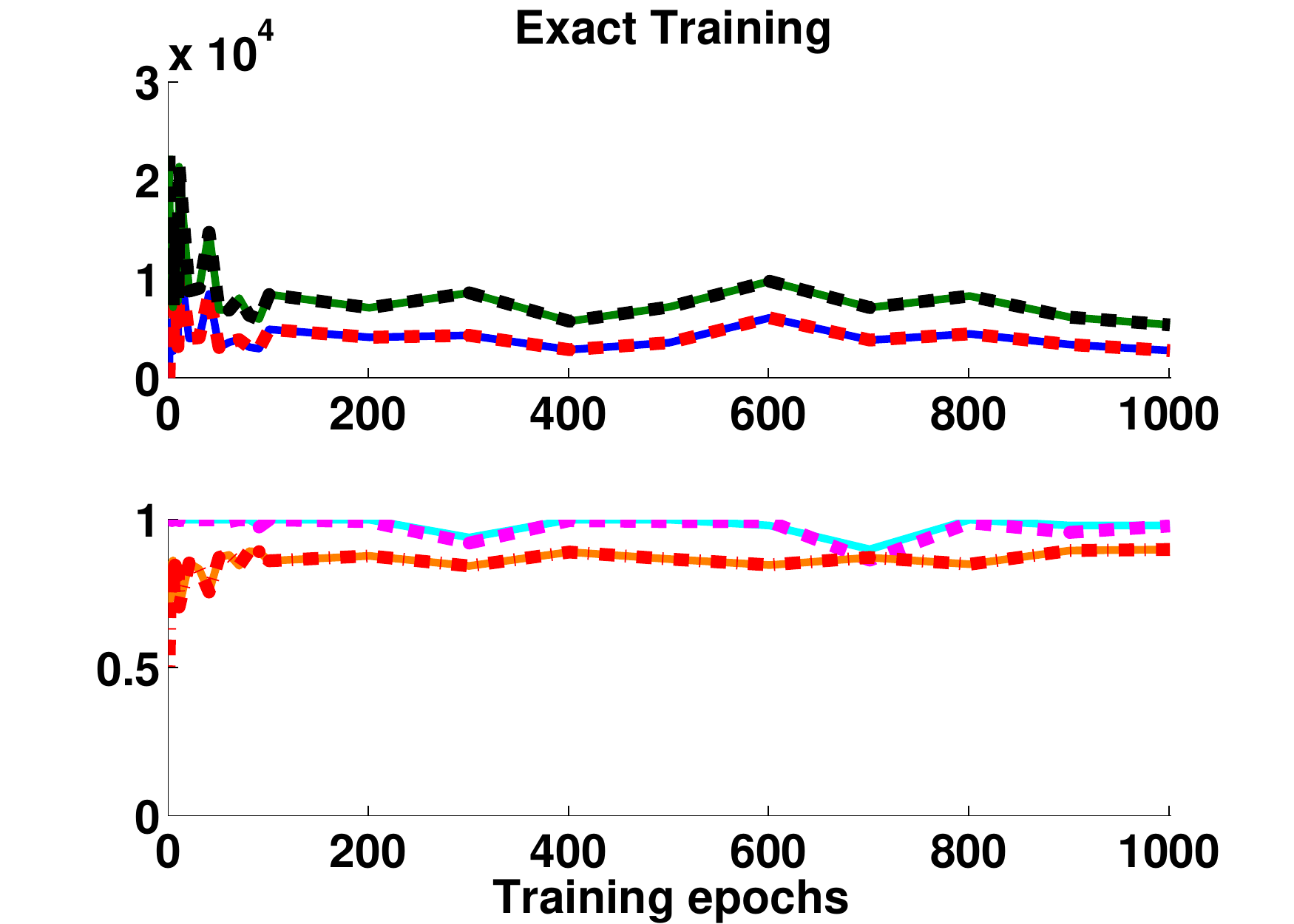}
  \end{tabular}
 \end{center}
\vspace{-4mm}
 \caption{{Training for foreground-background segmentation with the Weizmann Horse dataset. Various quantities of interest are shown as a function of training iterations. (Left) Training with LP relaxation. (Right) Training with ILP.}}
  \label{fig:horses}
\vspace{-0.4cm}
\end{figure*}

\paragraph{Image segmentation}
Finally, we conduct experiments on a foreground-background segmentation problem using the Weizmann Horse dataset \cite{borenstein2004}.
The data consists of 328 images, of which we use the first 50 for training and the rest for testing.
Here a binary output variable is assigned to each pixel, and there are $\sim58K$ variables per image on average.
We extract singleton and pairwise features as described in \citet{DomkePAMI2013}. 
\figref{fig:horses} 
shows the same quantities as in the multi-label setting, except for the accuracy measure -- here we compute the percentage of correctly classified pixels rather than $F_1$.
We observe a very similar behavior to that of the `Scene' multi-label dataset (\figref{fig:scene}).
Specifically, both relaxed and exact training produce a small integrality gap and high percentage of tight instances.
Unlike the `Scene' dataset, here only 1.2\% of variables satisfy the condition in Prop.~\ref{prop:strong_singles}  (using LP training). 
In all of our experiments the learned
model scores were never balanced (Prop.~\ref{prop:balanced}), although for the segmentation problem we believe the models learned are close to balanced, both for relaxed and exact training. 

\section{Conclusion}
\label{sec:conclusion}
In this paper we propose an explanation for the tightness of LP relaxations which has been observed in many structured prediction applications.
Our analysis is based on a careful examination of the integrality gap and its relation to the training objective. It shows how training with LP relaxations, although designed with accuracy considerations in mind, also induces tightness of the relaxation. 
Our derivation also suggests that exact training may sometimes have the opposite effect, increasing the integrality gap.

To explain tightness of test instances, we show that tightness generalizes from train to test instances.
Compared to the generalization bound of \citet{Kulesza07}, our bound only considers the tightness of the instance, ignoring label errors. Thus, for example, if learning happens to settle on a set of parameters in a tractable regime (e.g., supermodular potentials or stable instances \cite{MMVstability}) for which the LP relaxation is tight for all training instances, our generalization bound guarantees that with high probability the LP relaxation will also be tight on test instances. In contrast, in \citet{Kulesza07}'s bound, tightness on test instances can only be guaranteed when  the training data is algorithmically separable (i.e., LP-relaxed inference predicts perfectly). 


Our work suggests many directions for further study.
Our analysis in \secref{sec:train_integrality} focuses on the score hinge and ignores the task loss $\Delta$. It would be interesting to further study the effect of various task losses on tightness of the relaxation at training.
Next, our bound in \secref{sec:test_integrality} is intractable to compute due to the hardness of the surrogate loss $\varphi$. It is therefore desirable to derive a tractable alternative which could be used to obtain a useful guarantee in practice.
The upper bound on integrality shown in \secref{sec:train_integrality} holds for other convex relaxations which have been proposed for structured prediction, such as semi-definite programming relaxations \cite{KumarJMLR09}. 
However, it is less clear how to extend the generalization result to such non-polyhedral relaxations. 
Finally, we hope that our methodology will be useful for shedding light on tightness of convex relaxations in other learning problems.

\section*{Appendix}
\appendix

\section{$\gamma$-Tight LP Relaxations}
\label{app:gamma_tight}
In this section we provide full derivations for the results in \secref{sec:gamma_tight}.
We make extensive use of the results in \citet{weller16} (some of which are restated here for completeness).
We start by defining a model in minimal representation, which will be convenient for the derivations that follow.
Specifically, in the case of binary variables ($y_i\in\{0,1\}$) with pairwise factors, we define a value $\eta_i$ for each variable, and a value $\eta_{ij}$ for each pair. The mapping between the over-complete vector $\mu$ and the minimal vector $\eta$ is as follows.
For singleton factors, we have:
\[
\mu_i = \left( \begin{array}{c}
1-\eta_i \\ \eta_i
\end{array} \right)
\]
Similarly, for the pairwise factors, we have:
\[
\mu_{ij} = \left( \begin{array}{cc}
1 + \eta_{ij} - \eta_i - \eta_j & \eta_j - \eta_{ij}~, \\
\eta_i - \eta_{ij} & \eta_{ij}
\end{array} \right)
\]
The corresponding mapping to minimal parameters is then:
\begin{align*}
\bar\theta_i &= \theta_i(1) - \theta_i(0) + \sum_{j\in N_i} (\theta_{ij}(1,0) - \theta_{ij}(0,0)) \\
\bar\theta_{ij} &= \theta_{ij}(1,1) + \theta_{ij}(0,0) - \theta_{ij}(0,1) -  \theta_{ij}(1,0)
\end{align*}
In this representation, the LP relaxation is given by (up to constants):
\[
\max_{\eta\in{\mathbb{L}}} f(\eta) := \sum_{i=1}^n \bar\theta_i\eta_i + \sum_{ij\in{\cal{E}}} \bar\theta_{ij}\eta_{ij}
\]
where $\mathbb{L}$ is the appropriate transformation of $\localpoly$ to the equivalent reduced space of $\eta$:
\begin{align*}
0 &\le \eta_i \le 1 \qquad\qquad\qquad\;\; \forall i \\
\max(0,\eta_i+\eta_j-1) &\le \eta_{ij} \le \min(\eta_i,\eta_j) \qquad \forall ij\in {\cal{E}}
\end{align*}

If $\bar\theta_{ij}>0$ ($\bar\theta_{ij}<0$), then the edge is called \emph{attractive} (\emph{repulsive}).
If all edges are attractive, then the LP relaxation is known to be tight \cite{jordan}.
When not all edges are attractive, in some cases it is possible to make them attractive by \emph{flipping} a subset of the variables ($y_i\leftarrow 1-y_i$).%
\footnote{The flip-set, if exists, is easy to find by making a single pass over the graph (see \citet{weller15} for more details).}
In such cases the model is called \emph{balanced}.

In the sequel we will make use of the known fact that all vertices of the local polytope are half-integral (take values in $\{0,\half,1\}$) \cite{jordan}.
We are now ready to prove the propositions (restated here for convenience).

\subsection{Proof of Proposition \ref{prop:balanced}}
{\bf Proposition \ref{prop:balanced}}
{\it
A balanced model with a unique optimum is $(\alpha/2)$-tight, where $\alpha$ is the difference between the best and second-best (integral) solutions.
}
\begin{proof}
\citet{weller16} define for a given variable $i$ the function $F^i_{\mathbb{L}}(z)$, which returns for every $0\le z \le 1$ the constrained optimum:
\[
F^i_{\mathbb{L}}(z) = \max_{\substack{\eta\in\mathbb{L} \\ \eta_i=z}} f(\eta)
\]
Given this definition, they show that for a balanced model, $F^i_{\mathbb{L}}(z)$ is a \emph{linear function} \citep[][Theorem 6]{weller16}.

Let $m$ be the optimal score, let $\eta^1$ be the unique optimum integral vertex in minimal form so $f(\eta^1) = m$, and any other integral vertex has value at most $m-\alpha$.
Denote the state of $\eta^1$ at coordinate $i$ by $z^* = \eta^1_i$, and consider computing the constrained optimum holding $\eta_i$ to various states.
By assumption, any other integral vertex has value at most $m-\alpha$, therefore,
\begin{align*}
F_{\mathbb{L}}^i(z^*) =&~ m \\
F_{\mathbb{L}}^i(1-z^*) \le&~ m-\alpha
\end{align*}
(the second line holds with equality if there exists a second-best solution $\eta^2$ s.t.~$\eta^2_i\ne\eta^1_i$).
Since $F_{\mathbb{L}}^i(z)$ is a linear function, we have that:
\begin{equation}
\label{eq:frac_less_than_linear}
F_{\mathbb{L}}^i(1/2) \le m - \alpha/2
\end{equation}

Next, towards contradiction, suppose that there exists a fractional vertex $\eta^f$ with value $f(\eta^f) > m-\alpha/2$.
Let $j$ be a fractional coordinate, so $\eta^f_j = \half$ (since vertices are half-integral).
Our assumption implies that $F_{\mathbb{L}}^j(1/2) > m-\alpha/2$, but this contradicts \eqref{eq:frac_less_than_linear}.
Therefore, we conclude that any fractional solution has value at most $f(\eta^f) \le m-\alpha/2$.
\end{proof}

It is possible to check in polynomial time if a model is balanced, if it has a unique optimum, and compute $\alpha$.
This can be done by computing the difference in value to the second-best.
In order to find the second-best: one can constrain each variable in turn to differ from the state of the optimal solution, and recompute the MAP solution; finally, take the maximum over all these trials.

\subsection{Proof of Proposition \ref{prop:strong_singles}}
{\bf Proposition \ref{prop:strong_singles}}
{\it
If all variables satisfy the condition:
\vspace{-5pt}
\[
\bar\theta_i \ge -\sum_{j\in N_i^-} \bar\theta_{ij} + \beta
,\quad\text{or } \quad
\bar\theta_i \le -\sum_{j\in N_i^+} \bar\theta_{ij} - \beta
\vspace{-5pt}
\]
for some $\beta>0$, then the model is $(\beta/2)$-tight.
}
\begin{proof}
For any binary pairwise models, given singleton terms $\{\eta_i\}$, the \emph{optimal} edge terms are given by \citep[for details see][]{weller16}:
\[
\eta_{ij}(\eta_i,\eta_j) = \begin{cases}
\min(\eta_i,\eta_j) & \text{if } \bar\theta_{ij} > 0 \\
\max(0,\eta_i+\eta_j-1) & \text{if } \bar\theta_{ij} < 0
\end{cases}
\]
Now, consider a variable $i$ and let $N_i$ be the set of its neighbors in the graph. Further, define the sets $N_i^+=\{j\in N_i | \bar\theta_{ij}>0\}$ and $N_i^-=\{j\in N_i | \bar\theta_{ij}<0\}$, corresponding to attractive and repulsive edges, respectively.
We next focus on the parts of the objective affected by the value at $\eta_i$ (recomputing optimal edge terms); recall that all vertices are half-integral:
\begin{center}
\small{
\begin{tabular}{c|c|c}
$\eta_i = 1$ & $\eta_i = 1/2$ & $\eta_i = 0$ \\
\hline
\rule{0pt}{3ex}
$\bar\theta_i + \hspace{-7pt} \sum\limits_{\substack{j\in N_i^+ \\ \eta_j=1}} \hspace{-2pt} \bar\theta_{ij} + \half \hspace{-4pt} \sum\limits_{\substack{j\in N_i^+ \\ \eta_j=\half}} \hspace{-2pt} \bar\theta_{ij} + \hspace{-7pt} \sum\limits_{\substack{j\in N_i^- \\ \eta_j = 1}} \hspace{-3pt} \bar\theta_{ij} + \half \hspace{-5pt} \sum\limits_{\substack{j\in N_i^- \\ \eta_j = \half}} \hspace{-3pt} \bar\theta_{ij}$
&
$\half\bar\theta_i + \half \hspace{-9pt} \sum\limits_{\substack{j\in N_i^+ \\ \eta_j\in\{\half,1\}}} \hspace{-7pt} \bar\theta_{ij} + \half \hspace{-7pt} \sum\limits_{\substack{j\in N_i^- \\ \eta_j = 1}} \hspace{-4pt} \bar\theta_{ij}$
&
$0$
\end{tabular}
} 
\end{center}
It is easy to verify that the condition $\bar\theta_i \ge -\sum_{j\in N_i^-} \bar\theta_{ij} + \beta$ guarantees that $\eta_i=1$ in the optimal solution.
We next bound the difference in objective values resulting from setting $\eta_i=1/2$.
\begin{align*}
\Delta f = \half\left( \bar\theta_i + \sum_{\substack{j\in N_i^+ \\ \eta_j=1}} \bar\theta_{ij} + \sum_{\substack{j\in N_i^- \\ \eta_j \in\{\half,1\}}} \bar\theta_{ij}\right)
\ge
\half\left( \bar\theta_i + \sum_{j\in N_i^-} \bar\theta_{ij}\right)
\ge
\beta/2
\end{align*}

Similarly, when $\bar\theta_i \le -\sum_{j\in N_i^+} \bar\theta_{ij} - \beta$, then $\eta_i=0$ in any optimal solution.
The difference in objective values from setting $\eta_i=1/2$ in this case is:
\begin{align*}
\Delta f = - \half \left( \bar\theta_i + \sum_{\substack{j\in N_i^+ \\ \eta_j\in\{\half,1\}}} \bar\theta_{ij} + \sum_{\substack{j\in N_i^- \\ \eta_j = 1}} \bar\theta_{ij} \right)
\ge
- \half \left( \bar\theta_i + \sum_{j\in N_i^+} \bar\theta_{ij} \right)
\ge
\beta/2
\end{align*}

Notice that for more fractional coordinates the difference in values can only increase, so in any case the fractional solution is worse by at least $\beta/2$.
\end{proof}


\begin{figure*}[t]
 \begin{center}
  \begin{tabular}{c c c}
  \hspace{-30px}
  \includegraphics[height=1.7in]{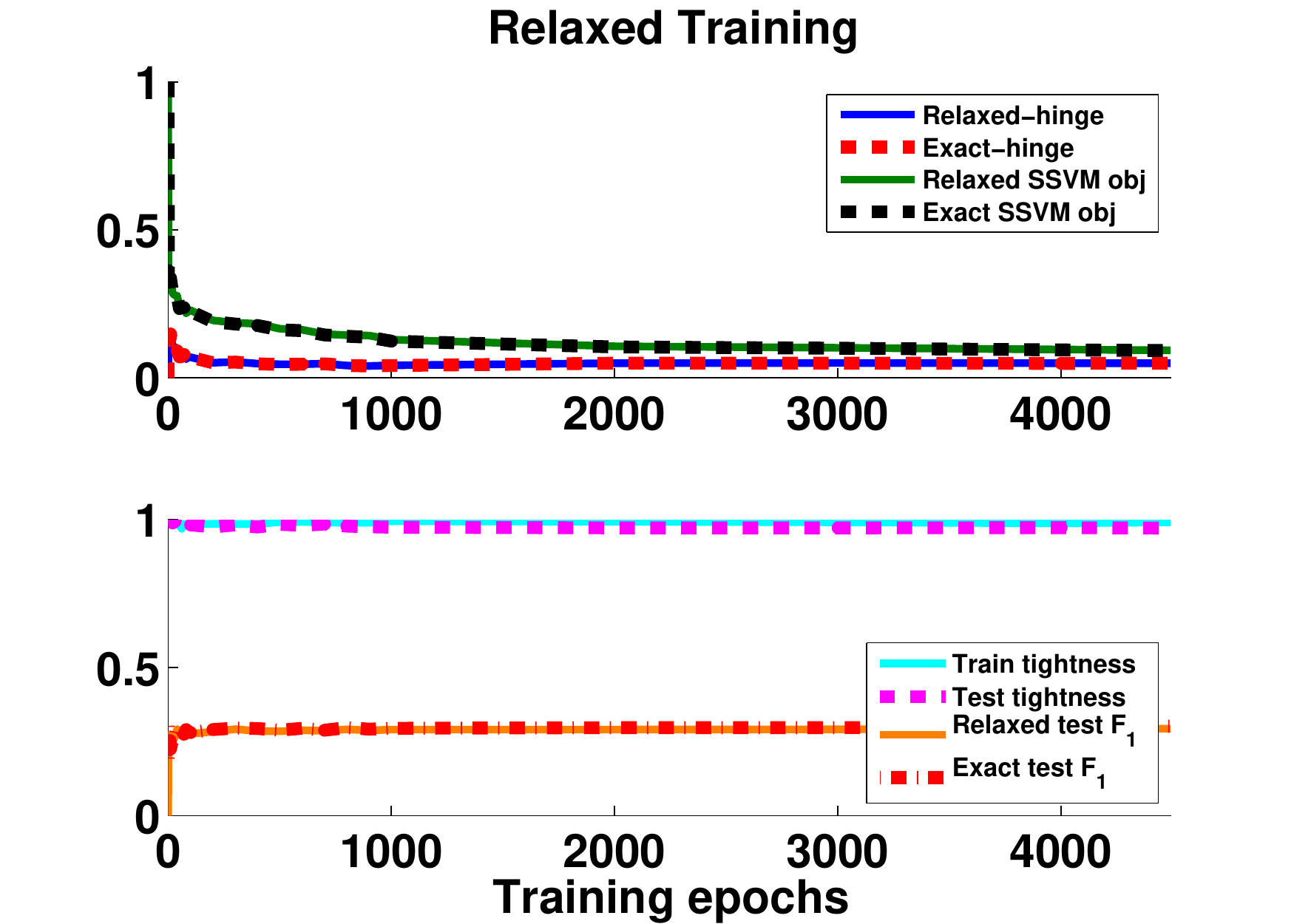}
  &
  \hspace{-25px}
  \includegraphics[height=1.7in]{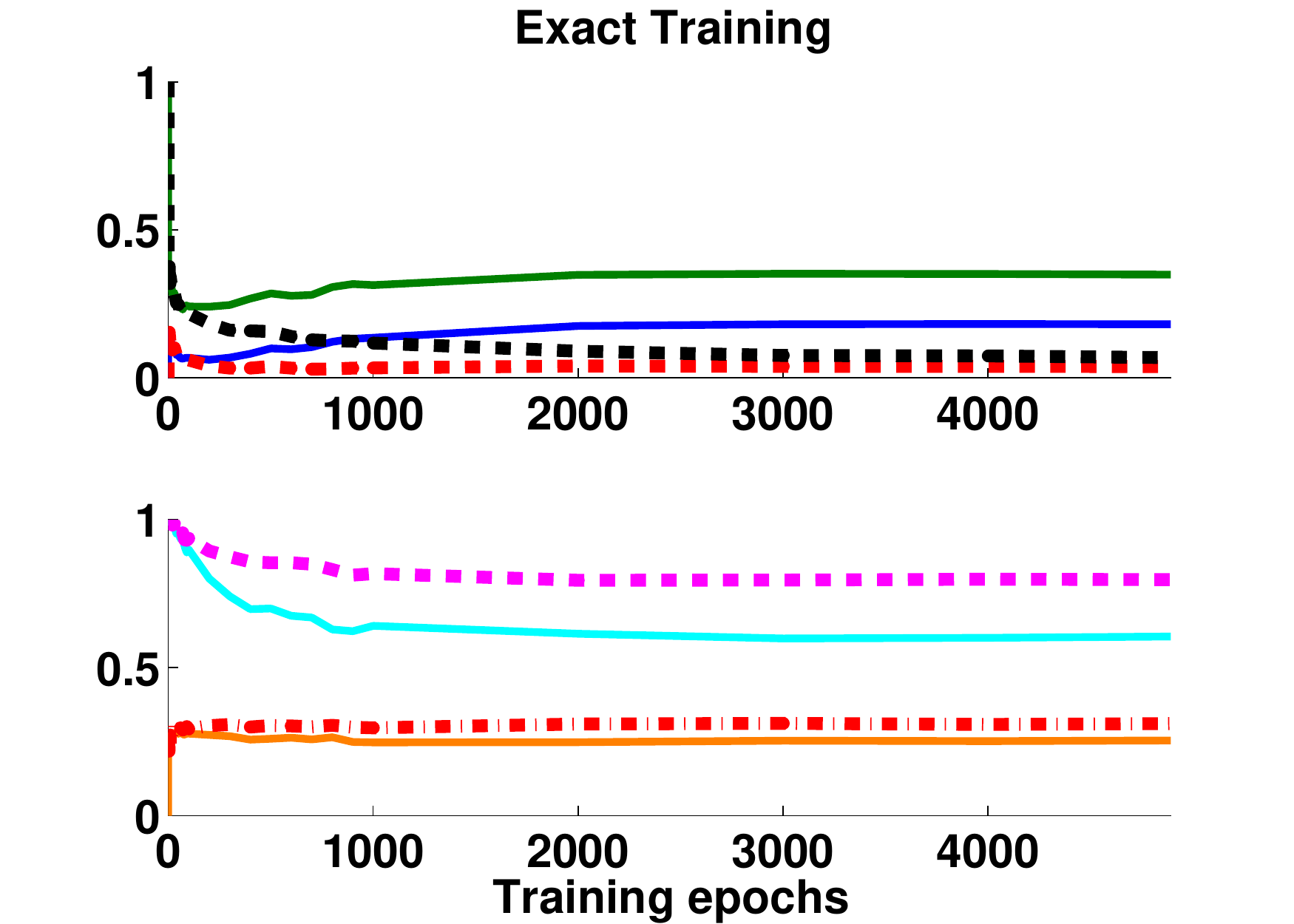}
  &
  \hspace{-10px}
  \includegraphics[height=1.7in]{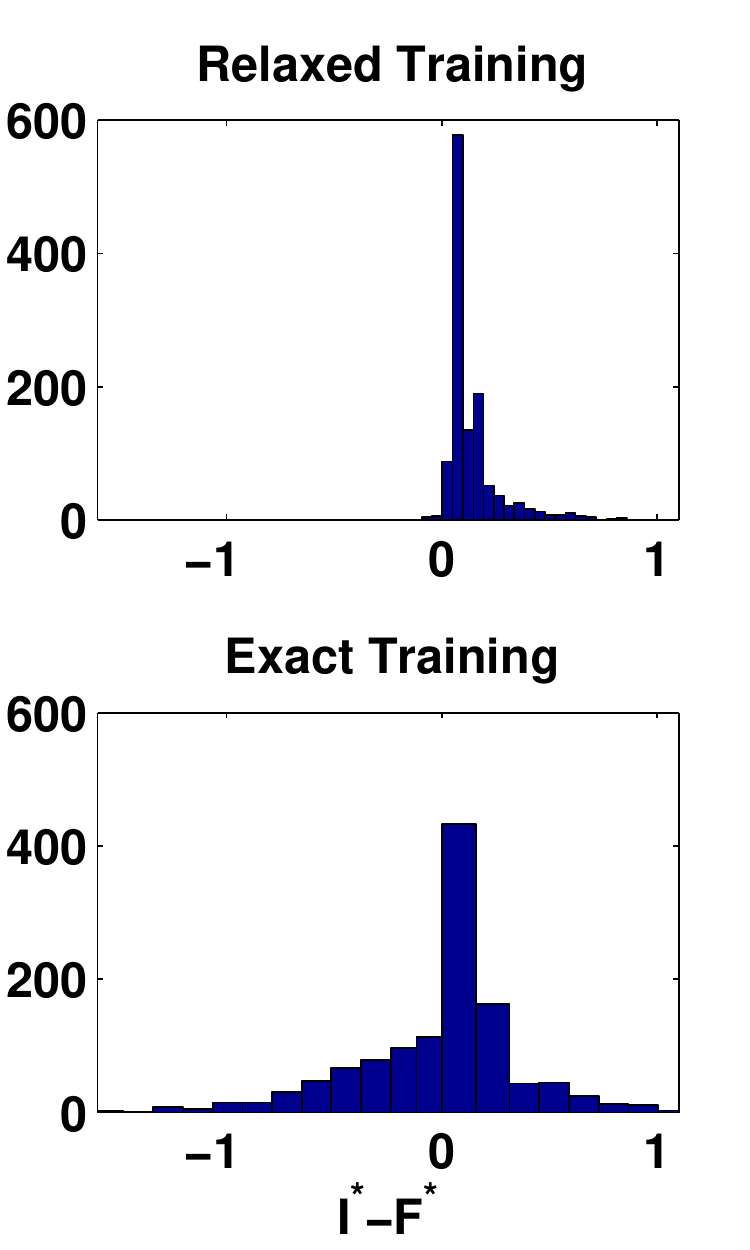}
  \end{tabular}
 \end{center}
\vspace{-4mm}
 \caption{{Training with a noisy version of the `Scene' dataset. Various quantities of interest are shown as a function of training iterations. (Left) Training with LP relaxation. (Middle) Training with ILP. (Right) Integrality margin (bin widths are scaled differently).}}
  \label{fig:scene_noisy}
\end{figure*}

\section{Additional Experimental Results}
\label{app:scene_noisy}
In this section we present additional experimental results for the `Scene' dataset.
Specifically, we inject random Gaussian noise to the input features in order to reduce the signal in the singleton scores and increase the role of the pairwise interactions.
This makes the problem harder since the prediction needs to account for global information.

In \figref{fig:scene_noisy} we observe that with exact training the exact loss is minimized, causing the exact-hinge to decrease, since it is upper bounded by the loss (middle, top). On the other hand, the relaxed-hinge (and relaxed loss) \emph{increase} during training, which results in a large integrality gap and fewer tight instances.
In contrast, with relaxed training the relaxed loss is minimized, which causes the relaxed-hinge to decrease. Since the exact-hinge is upper bounded by the relaxed-hinge it also decreases, but both hinge terms decrease similarly and remain very close to each other.
This results in a small integrality gap and tightness of almost all instances.

Finally, in contrast to other settings, in \figref{fig:scene_noisy} we observe that with exact training the test tightness is noticeably higher (about 20\%) than the train tightness (\figref{fig:scene_noisy}, middle, bottom).
This does not contradict our bound from \thmref{thm:integrality_generalization_bound}, since in fact the test fractionality is even \emph{lower} than the bound suggests.
On the other hand, this result does entail that train and test tightness may sometimes behave differently, which means that we might need to increase the size of the trainset in order to get a tighter bound. 


\clearpage
\bibliographystyle{abbrvnat}
\begin{spacing}{0.5}
\small{
\bibliography{my_bib}
}
\end{spacing}

\end{document}